# Parameter-Free Spectral Kernel Learning

**Qi Mao**
School of Computer Engineering
Nanyang Technological University
Singapore, 639798

**Ivor W. Tsang**
School of Computer Engineering
Nanyang Technological University
Singapore, 639798

## Abstract

Due to the growing ubiquity of unlabeled data, learning with unlabeled data is attracting increasing attention in machine learning. In this paper, we propose a novel semi-supervised kernel learning method which can seamlessly combine manifold structure of unlabeled data and Regularized Least-Squares (RLS) to learn a new kernel. Interestingly, the new kernel matrix can be obtained analytically with the use of spectral decomposition of graph Laplacian matrix. Hence, the proposed algorithm does not require any numerical optimization solvers. Moreover, by maximizing kernel target alignment on labeled data, we can also learn model parameters automatically with a closed-form solution. For a given graph Laplacian matrix, our proposed method does not need to tune any model parameter including the trade-off parameter in RLS and the balance parameter for unlabeled data. Extensive experiments on ten benchmark datasets show that our proposed two-stage parameter-free spectral kernel learning algorithm can obtain comparable performance with fine-tuned manifold regularization methods in transductive setting, and outperform multiple kernel learning in supervised setting.

## 1 Introduction

Experimental results show that the performance of kernel based machine learning methods is usually determined by the kernel function rather than the kernel machine. Therefore, the choice of an effective kernel is primarily important for kernel methods. Traditionally, kernel methods, such as Support Vector Machine (SVM) often adopt a predefined kernel chosen from a pool of parametric kernel functions (e.g. Gaussian and polynomial kernels). The limitation of this approach lies in the difficulty of tuning optimal kernel parameters for the predefined parametric kernel functions. To address this issue, learning effective kernels from data automatically has been actively explored in machine learning community for several years.

Multiple Kernel Learning (MKL) is one major stream of kernel learning proposed by Lanckriet et al. (2004), which learns a new kernel from a linear combination of predefined base kernels. It has been actively studied (Bach et al., 2004; Sonnenburg et al., 2006; Rakotomamonjy et al., 2008) and achieved successful results in many applications including computer vision (Duan et al., 2009; Sun et al., 2009; Vedaldi et al., 2009). In addition, there are other MKL methods along this line, such as non-linear combination of predefined base kernels (Cortes et al., 2009b), variety of norm regularizer (Cortes et al., 2009a; Kloft et al., 2009; Nath et al., 2009). These methods are usually designed for kernel methods in supervised learning setting.

Spectral Kernel Learning (SKL) is another popular direction of kernel learning methods. The basic idea is to learn a spectral structure of the new kernel from a given kernel/similarity matrix. These methods include parametric kernel methods, such as diffusion kernels in supervised setting (Kondor and Lafferty, 2002), cluster kernels (Chapelle et al., 2002) and Gaussian random field kernels (Zhu et al., 2003) in semi-supervised setting. These methods usually assume certain parametric kernel forms, which limit their capacity of fitting diverse patterns in real applications. To avoid this limitation, Zhu et al. (2004) proposed to incorporate ordered constraints into SKL.

Besides these, several kernel learning methods (Kwok and Tsang, 2003; Hoi et al., 2007; Liu et al., 2009; Zhuang et al., 2009) have been proposed in semi-supervised setting by using side information. Note, most of these kernel learning methods rely on advanced numerical optimization solvers, such as Semi-Definite Programming (SDP) (Lanckriet et al., 2004), Quadratic Constrained Quadratic Programming (QCQP) (Zhu et al., 2004), Semi-Infinite Linear Programming (SILP) (Sonnenburg et al., 2006), Linear Programming (LP) (Liu et al., 2009).

In this paper, we study kernel learning for Regularized Least-Squares (RLS), as it can be easily interpreted as a Bayesian model which can give probabilistic output for prediction. RLS has also been successfully used in both kernel learning and Semi-Supervised Learning (SSL). Moreover, Cortes et al. (2009a,b) recently demonstrated several statistical properties of RLS that lead to successful results in MKL. Manifold regularization approach such as Laplacian RLS (LapRLS) has been shown achieving state-of-the-art performance in semi-supervised setting (Belkin et al., 2006), and it is more robust than Laplacian SVM (LapSVM) (Chapelle et al., 2006). However, manifold regularization approach typically introduces many model parameters which make the model selection of kernel parameters more complicated.

The key observation is that kernel learning or semi-supervised learning problems often need to introduce more parameters so as to build a model upon non-parametric prior information, such as graph Laplacian matrix or similarity matrix. Moreover, the generalization performance also critically depends on model parameters. With limited labeled data, it may be ineffective to learn model parameters by cross-validation or Bayesian approaches from labeled data only. Therefore, how to learn model parameters effectively and efficiently is still a very challenging task.

In this paper, we propose a novel semi-supervised kernel learning method which can seamlessly combine manifold structure of unlabeled data and RLS to learn a new kernel. The major contributions of our proposed method include:

1. Unlike previous methods, our proposed method does not need any numerical optimization solvers. The learned non-parametric kernel matrix can be obtained analytically with the use of spectral decomposition of graph Laplacian matrix.

2. We can also automatically learn model parameters defined in the framework with a closed-form solution by maximizing kernel target alignment on labeled data.

3. Moreover, it can be shown that our proposed two-stage SKL algorithm does not contain any parameter if given a specific graph Laplacian matrix ahead of time. Similar to Bayesian inference framework for ARD kernel in Gaussian Process, we do not require to perform any model selection by validation methods.

4. Furthermore, our proposed algorithm can learn a new kernel and the decision function simultaneously from both labeled and unlabeled data.

5. The classification performance of our proposed method on ten benchmark datasets is significantly better than MKL algorithms in supervised setting, and comparable with the fine-tuned manifold regularization methods using expensive cross-validation in transductive setting.

In the sequel, the transpose of a matrix/vector is denoted by a superscript $T$, and tr($A$) denotes the trace of matrix $A$. $A \succ \mathbf{0}$ (resp. $A \succeq \mathbf{0}$) means matrix $A$ is symmetric and positive definite (pd) (resp. positive semidefinite (psd)). The identity matrix is denoted by $I$.

## 2 Multiple Kernel Learning for RLS

Given a labeled training dataset $\mathbb{D}_l = \{(x_i, y_i)\}_{i=1}^{n_l}$ where $x_i$ is the input and $y_i$ is the output of the $i$th pattern, the decision function $f(x) = w^T \phi(x)$ of RLS can be learned by minimizing the following regularized risk functional:

$$\min_w \frac{1}{2}||w||^2 + \frac{C}{2}\sum_{i=1}^{n_l}(y_i - w^T\phi(x_i))^2, \quad (1)$$

where $C > 0$ is the tradeoff parameter to regulate model complexity $||w||^2$ and empirical loss of RLS, and $\phi(\cdot)$ is a nonlinear feature map of $x_i$. By introducing slack variables $\xi_i$'s to form equality constraints $\xi_i = y_i - w^T\phi(x_i), \forall i = 1, \ldots, n_l$, and the corresponding dual variable $\alpha_i$ for each equality constraint, the minimization problem (1) can be derived into its dual form as follows:

$$\max_\alpha \alpha^T \mathbf{y}_l - \frac{1}{2}\alpha^T\Big(K_{l,l} + \frac{1}{C}I_l\Big)\alpha, \quad (2)$$

where $\alpha = [\alpha_1, \ldots, \alpha_{n_l}]^T$ is a vector of dual variables, $\mathbf{y}_l = [y_1, \ldots, y_{n_l}]^T$ is a vector of labels, $K_{l,l} \succeq \mathbf{0}$ is a $n_l \times n_l$ kernel matrix with entries $k(x_i, x_j) = \langle \phi(x_i), \phi(x_j) \rangle_\mathcal{H}$ representing an inner product of $\phi(x_i)$ and $\phi(x_j)$ in a Reproducing Kernel Hilbert Space (RKHS) $\mathcal{H}$, and the primal variable $w$ in (1) can be recovered by $w = \sum_{i=1}^{n_l} \alpha_i \phi(x_i)$. Moreover, by setting the derivatives of (2) w.r.t. $\alpha$ to zeros, we obtain a closed-form solution

$$\alpha = \Big(K_{l,l} + \frac{1}{C}I_l\Big)^{-1}\mathbf{y}_l. \quad (3)$$

In MKL, the kernel is assumed to be a linear combination of $m$ predefined base kernels, i.e., $K_{l,l} = \sum_{d=1}^{m} \mu_d K_{l,l}^d$, where $\mu_d$ is the coefficient for the $d$th base kernel matrix $K_{l,l}^d$, then the MKL problem for RLS can be written as:

$$\min_{\mu_d \geq 0} \max_\alpha \alpha^T \mathbf{y}_l - \frac{1}{2}\alpha^T\Big(\sum_{d=1}^m \mu_d K_{l,l}^d + \frac{1}{C}I_l\Big)\alpha, \quad (4)$$

which can be reformulated as a QCQP problem. Recently, Cortes et al. (2009a) proposed an iterative algorithm for (4), which update the coefficients $\mu_d$'s and the dual variables $\alpha$ alternatively. In particular, $\mu_d$ is updated by gradient descent and $\alpha$ is obtained by (3).

## 3 Manifold Regularization for Kernel Learning

As discussed in manifold regularization (Belkin et al., 2006), the intrinsic geometric structure of data can be explored to learn a better decision function effectively, particularly, when a lot of unlabeled data are readily available. Note, however the MKL formulation in (4) does not consider the use of unlabeled data. Several kernel learning approaches (Hoi et al., 2007; Zhuang et al., 2009) have studied to adopt the data manifold (Belkin et al., 2006) for preserving the locality in kernel learning.

Given a set of unlabeled data $\mathbb{D}_u = \{x_i\}_{i=1+n_l}^{n_l+n_u}$, and $\mathbb{D} = \mathbb{D}_l \cup \mathbb{D}_u$ with $n = n_l + n_u$ patterns, $S$ is a $n \times n$ similarity matrix with entries $S_{i,j} \geq 0$, which defines a similarity measure between any two patterns $x_i$ and $x_j \in \mathbb{D}$, and $K \succeq 0$ a $n \times n$ kernel matrix defined on $\mathbb{D}$. Assume that this kernel matrix can be expressed as $K = V^T V$ where $V = [v_1, \ldots, v_n]^T$ is a matrix of the embedding of patterns. A manifold regularizer which captures the local dependency between the embedding of $v_i$ and $v_j$, can be written as:

$$\sum_{i,j=1}^{n} S_{i,j} \left\| \frac{v_i}{\sqrt{d_i}} - \frac{v_j}{\sqrt{d_j}} \right\|_2^2 = \mathrm{tr}(VLV^T)$$
$$= \mathrm{tr}(KL), \quad (5)$$

where $L$ is the graph Laplacian matrix defined as: $L = I - D^{-1/2} S D^{-1/2}$, where $D$ is a diagonal matrix with the diagonal entries defined as $d_i = D_{i,i} = \sum_{j=1}^{N} S_{ij}$.

## 4 Parameter-Free Spectral Kernel Learning

In this Section, we introduce our proposed two-stage SKL framework for RLS. Recall that when we substitute (3) back into the dual (2), the optimal objective value is equal to

$$\frac{1}{2} \mathbf{y}_l^T \left( K_{l,l} + \frac{1}{C} I_l \right)^{-1} \mathbf{y}_l.$$

Together with the manifold regularizer in (5), we introduce a semi-supervised kernel learning problem as follows:

$$\min_{K \succeq 0} \frac{1}{2} \mathbf{y}_l^T \left( K_{l,l} + \frac{1}{C} I_l \right)^{-1} \mathbf{y}_l + \mu \mathrm{tr}(KL), \quad (6)$$

where $\mu > 0$ regulates the two terms.

Following the trick used in (Lanckriet et al., 2004), one can transform (6) into the following optimization problem,

$$\min_{\tau, K \succeq 0} \quad \tau + \mu \mathrm{tr}(KL)$$
$$\text{subject to} \quad \begin{bmatrix} K_{l,l} + \frac{1}{C} I_l & \mathbf{y}_l \\ \mathbf{y}_l^T & \tau \end{bmatrix} \succeq \mathbf{0},$$

which is in form of a Semi-Definite Programming (SDP) problem, and it is computationally expensive even when the size of $\mathbb{D}$ is small.

In order to solve Problem (6), we give an upper bound of this problem in Proposition 1.

**Proposition 1.** *Denote* $\mathbf{y} = \begin{bmatrix} \mathbf{y}_l \\ \mathbf{0} \end{bmatrix}$, *the optimization problem*

$$\min_{K \succeq \mathbf{0}} \frac{1}{2} \mathbf{y}^T \left( K + \frac{1}{C} I \right)^{-1} \mathbf{y} + \mu tr(KL), \quad (7)$$

*is an upper bound of Problem (6), and which is the same as using 0 as the label for unlabeled data in Problem (6).*

*Proof.* According to matrix inversion in block form, given matrices $\begin{bmatrix} \mathcal{A} & \mathcal{B} \\ \mathcal{E} & \mathcal{Q} \end{bmatrix}^{-1}$ where $\mathcal{A}$ and $\mathcal{Q}$ are invertible, it is equivalent to

$$\begin{bmatrix} (\mathcal{A} - \mathcal{B}\mathcal{Q}^{-1}\mathcal{E})^{-1} & -\mathcal{A}^{-1}\mathcal{B}(\mathcal{Q} - \mathcal{E}\mathcal{A}^{-1}\mathcal{B})^{-1} \\ -\mathcal{Q}^{-1}\mathcal{E}(\mathcal{A} - \mathcal{B}\mathcal{Q}^{-1}\mathcal{E})^{-1} & (\mathcal{Q} - \mathcal{E}\mathcal{A}^{-1}\mathcal{B})^{-1} \end{bmatrix}.$$

For the same definition of all matrices, the Woodbury formula can be formulated as follows:

$$(\mathcal{A} - \mathcal{B}\mathcal{Q}^{-1}\mathcal{E})^{-1} = \mathcal{A}^{-1} + \mathcal{A}^{-1}\mathcal{B}(\mathcal{Q} - \mathcal{E}\mathcal{A}^{-1}\mathcal{B})^{-1}\mathcal{E}\mathcal{A}^{-1}.$$

Assume $\mathcal{X} = \mathcal{X}^T = \begin{bmatrix} \mathcal{A} & \mathcal{B} \\ \mathcal{B}^T & \mathcal{Q} \end{bmatrix}$ and $\mathcal{A}$ is symmetric matrix. The Schur complement lemma states that if $\mathcal{A} \succ \mathbf{0}$, then $\mathcal{X} \succeq \mathbf{0}$ if and only if $\mathcal{Q} - \mathcal{B}^T \mathcal{A}^{-1} \mathcal{B} \succeq \mathbf{0}$.

For convenience, we denote $\mathcal{A} = K + \frac{1}{C} I \succ \mathbf{0}$ since $K \succeq \mathbf{0}$. We assume that $\mathcal{A} = \begin{bmatrix} \mathcal{A}_{l,l} & \mathcal{B} \\ \mathcal{B}^T & \mathcal{Q} \end{bmatrix} \succeq \mathbf{0}$, given some matrix $\mathcal{B}$ and invertible $\mathcal{Q}$, according to above equations, we can obtain the derivative as follows:

$$\mathbf{y}^T \mathcal{A}^{-1} \mathbf{y} = [\mathbf{y}_l^T \; \mathbf{0}^T] \begin{bmatrix} \mathcal{A}_{l,l} & \mathcal{B} \\ \mathcal{B}^T & \mathcal{Q} \end{bmatrix}^{-1} \begin{bmatrix} \mathbf{y}_l \\ \mathbf{0} \end{bmatrix}$$
$$= \mathbf{y}_l^T (\mathcal{A}_{l,l} - \mathcal{B}\mathcal{Q}^{-1}\mathcal{B}^T)^{-1} \mathbf{y}_l$$
$$= \mathbf{y}_l^T (\mathcal{A}_{l,l}^{-1} + \mathcal{M}) \mathbf{y}_l,$$

where $\mathcal{M} = \mathcal{A}_{l,l}^{-1} \mathcal{B} (\mathcal{Q} - \mathcal{B}^T \mathcal{A}_{l,l}^{-1} \mathcal{B})^{-1} \mathcal{B}^T \mathcal{A}_{l,l}^{-1}$. According to the Schur complement lemma, we can obtain $\mathcal{M} \succeq \mathbf{0}$, and then $\mathbf{y}_l^T \mathcal{M} \mathbf{y}_l \geq 0$. Therefore, the following inequality can be obtained,

$$\mathbf{y}^T \left( K + \frac{1}{C} I \right)^{-1} \mathbf{y} \geq \mathbf{y}_l^T \left( K_{l,l} + \frac{1}{C} I_l \right)^{-1} \mathbf{y}_l.$$

This completes the proof. $\square$

This upper bound is similar to Learning with Local and Global Consistency (Zhou et al., 2003) for semi-supervised learning, which uses 0 as the label of unlabeled data. In this case, there is no relaxation. Problem (7) also can be reformulated as a SDP problem. To alleviate the computational burden, in the next subsections, we apply SKL techniques to simplify the SDP constraint $K \succeq \mathbf{0}$, and employ kernel target alignment to learn parameter $\mu$.

### 4.1 A Closed-Form Solution via Spectral Kernel

In SKL, the kernel matrix $K$ is learned from the eigenvectors of a given kernel or a graph Laplacian matrix. Moreover, the positive semi-definite property of $K$ can be determined automatically if the spectral can be constrained as nonnegative values. Specifically, given a graph Laplacian matrix $L$ such that $L = U \Upsilon U^T$, where $\Upsilon$ is a diagonal matrix with diagonal entries equal to eigenvalues $\gamma_i$'s of $L$ and $U$ is a matrix of column eigenvectors of $L$, the new kernel can be formulated as $K = U \Sigma U^T$ based on $\mathbb{D}$. Here, we assume $\Sigma$ is a diagonal matrix such that $\Sigma = \mathrm{diag}(\Lambda)$ and $\Lambda = [\lambda_1, \ldots, \lambda_n]$ is a vector of variables $\lambda_i$'s ($\lambda_i \geq 0$).

The following proposition shows that Problem (7) could be solved in the closed form solution via spectral kernel.

**Proposition 2.** *If $K$ is represented as a spectral kernel $K = U\Sigma U^T$, where $U$ is a matrix of column eigenvectors of $L$, $\Sigma = diag(\Lambda)$, and $\Lambda = [\lambda_1, \ldots, \lambda_n]$ and $\lambda_i \geq 0$,*

*Problem (7) has the closed-form solution $\lambda^*$:*

$$\lambda_i^* = \left[\sqrt{\frac{A_{i,i}}{2\mu B_{i,i}}} - \frac{1}{C}\right]_+, \forall i = 1, \ldots, n, \quad (8)$$

*where* $[u]_+ = \max(0, u)$, $A = U_l^T \mathbf{y}_l \mathbf{y}_l^T U_l$, $B = U^T L U = diag([\gamma_1, \ldots, \gamma_n])$ *and $U_l$ contains the first $n_l$ rows of $U$.*

*Proof.* Denote the objective of (7) by $F(\Lambda)$. We arrive at the problem to minimize the following objective:

$$F(\Lambda) = \frac{1}{2} tr\left(\mathbf{y}^T \left(U\Sigma U^T + \frac{1}{C}I\right)^{-1} \mathbf{y}\right) + \mu tr(U\Sigma U^T L).$$

According to the property of eigenvectors, $U^T U = I$, then $F(\Lambda)$ can be further derived as follows:

$$\frac{1}{2} tr(\mathbf{y}^T (U\Sigma U^T + \frac{1}{C}I)^{-1}\mathbf{y}) + \mu tr(\Sigma \Upsilon)$$
$$= \frac{1}{2} tr(\mathbf{y}^T U(\Sigma + \frac{1}{C}I)^{-1}U^T \mathbf{y}) + \mu tr(\Sigma \Upsilon)$$
$$= \frac{1}{2} tr(\mathbf{y}_l^T U_l(\Sigma + \frac{1}{C}I)^{-1}U_l^T \mathbf{y}_l) + \mu tr(\Sigma \Upsilon)$$
$$= \frac{1}{2}\sum_{i=1}^n \frac{1}{\lambda_i + \frac{1}{C}} A_{i,i} + \mu \sum_{i=1}^n \lambda_i B_{i,i},$$

where $U_l$ contains the first $n_l$ rows of $U$, $A = U_l^T \mathbf{y}_l \mathbf{y}_l^T U_l$ and $B = U^T L U = \Upsilon$. By setting the derivatives of $F(\lambda)$ to zeros, we get the analytical solution of Problem (7):

$$\lambda_i = \sqrt{\frac{A_{i,i}}{2\mu B_{i,i}}} - \frac{1}{C} = \sqrt{\frac{A_{i,i}}{2\mu \gamma_i}} - \frac{1}{C}, \forall i = 1, \ldots, n.$$

Since $\lambda_i$ should be non-negative, the projection to non-negative value for each $\lambda_i$ is needed. □

### 4.2 Learning $\mu$ using Kernel Target Alignment

The analytical solution (8) makes (7) easier to be solved. However, this solution includes two free parameters $\mu$ and $C$. So, the final performance greatly depends on the choice of these two parameters. Since $\mu$ and $C$ strongly depends on each other, cross-validation by grid search method could not be efficient because $C$ and $\mu$ directly affect the eigenvalues of $K$. Moreover, cross-validation may not be effective with very few labeled data. Here, we propose to learn the parameter $\mu$ automatically so as to avoid above limitations by maximizing kernel target alignment criterion.

Note, solution (8) can be regarded as one special case for SKL. In SKL, the new kernel matrix $K$ is formed as $K = \sum_{i=1}^r \lambda(\gamma_i) \mathbf{v}_i \mathbf{v}_i^T$, where $\{(\mathbf{v}_i, \gamma_i)\}$ is the eigensystem from a given graph Laplacian matrix $L$, and $\lambda(\gamma_i) = \lambda_i$. Zhu et al. (2004); Smola and Kondor (2003) suggest that function $\lambda(.)$ reverses the order of the eigenvalues of $L$, so that a smoothing $\mathbf{v}_i$ has a lower penalty. Since regularization term is a function of $\lambda(\gamma)$, and $\lambda$ is decreasing, a larger penalty is incurred for those functions corresponding to eigenfunctions that are less smooth. Apparently, solution (8) exactly satisfies the need of above graph theory. Our solution is still different from the other parametric forms, such as diffusion kernel (Kondor and Lafferty, 2002) $\lambda(\gamma_i) = \exp(-\frac{\sigma^2}{2}\gamma_i)$ and Gaussian field kernel (Zhu et al., 2003) $\lambda(\gamma_i) = \frac{1}{\gamma_i + \epsilon}$, since our function integrates the label information indicated by $A$ and the manifold structure encoded by $B$.

Similar to (Zhu et al., 2004), kernel target alignment criterion (KTA) can be used to determine the new kernel. Instead of learning $\lambda_i$, we use KTA to learn parameter $\mu$. The KTA (Cristianini et al., 2002) can be written as:

$$KTA(K, \mathbf{yy}^T) = \frac{\langle K, \mathbf{yy}^T \rangle_F}{\sqrt{\langle K, K^T \rangle_F \langle \mathbf{yy}^T, \mathbf{yy}^T \rangle_F}}$$
$$= \frac{tr(\Sigma A)}{\sqrt{tr(\Sigma^2)} n_l^2},$$

where $\langle M, N \rangle_F$ denotes the Frobenius product between two square matrices such that $\langle M, N \rangle_F = \sum_{i,j} M_{i,j} N_{i,j} = tr(MN^T)$, and the last equality holds due to $K = U\Sigma U^T$ and $A = U_l^T \mathbf{y}_l \mathbf{y}_l^T U_l$ in Section 4.1.

According to (Cristianini et al., 2002), the larger the KTA is, the closer the learned kernel approaches to the ideal kernel $\mathbf{yy}^T$. Intuitively, the parameter $\mu$ should be learned by maximizing KTA criterion as follows:

$$\max_{\mu > 0} KTA(K, \mathbf{yy}^T). \quad (9)$$

The following Proposition shows that Problem (9) could be solved analytically.

**Proposition 3.** *By maximizing the KTA for $K$ in Proposition 2, Problem (9) has the closed form solution $\mu^*$:*

$$\mu^* = \left(\frac{\overline{yu} - \overline{zx}}{\overline{zu} - \frac{n}{C^2}\overline{x}}\right)^2, \quad (10)$$

*where* $\overline{x} = \sum_{i=1}^n A_{i,i}\sqrt{\frac{A_{i,i}}{2B_{i,i}}}$, $\overline{y} = \sum_{i=1}^n \frac{A_{i,i}}{2B_{i,i}}$, $\overline{z} = \frac{1}{C}\sum_{i=1}^n \sqrt{\frac{A_{i,i}}{2B_{i,i}}}$ *and* $\overline{u} = \frac{1}{C}\sum_{i=1}^n A_{i,i}$.

*Proof.* By substituting $\Sigma$ with (8), (9) can be rewritten as:

$$\max_{\mu>0} \sum_{i=1}^n A_{i,i}\left(\sqrt{\frac{A_{i,i}}{2\mu B_{i,i}}} - \frac{1}{C}\right) / \sqrt{\sum_{i=1}^n \left(\sqrt{\frac{A_{i,i}}{2\mu B_{i,i}}} - \frac{1}{C}\right)^2}.$$

Since $K \succeq 0$, $tr(K\mathbf{yy}^T) = \mathbf{y}^T K \mathbf{y} \geq 0$ indicates that the value of KTA is always larger than or equal to zero. It also cannot be zero because neither $K$ nor $y$ is zero matrix or zero vector in the semi-supervised setting. Given a function $g(x) > 0$ for each variable $x \in R^d$, the optimal solution $x^* = \arg\max_x g(x) = \arg\max_x \log g(x)$.

By taking the logarithmic function of above objective, we can get the following equivalent optimization problem: $\max_{\mu>0} g(\mu)$, where

$$g(\mu) = \log \sum_{i=1}^n A_{i,i}\left(\sqrt{\frac{A_{i,i}}{2\mu B_{i,i}}} - \frac{1}{C}\right) - \frac{1}{2}\log \sum_{i=1}^n \left(\sqrt{\frac{A_{i,i}}{2\mu B_{i,i}}} - \frac{1}{C}\right)^2.$$

The derivative of $g(\mu)$ w.r.t. $\mu$ is shown as follows:

$$\nabla_\mu g(\mu) = \left(-\frac{1}{2}\mu^{-\frac{3}{2}}\right) \frac{\sum_{i=1}^n A_{i,i}\sqrt{\frac{A_{i,i}}{2B_{i,i}}}}{\sum_{i=1}^n A_{i,i}\left(\sqrt{\frac{A_{i,i}}{2\mu B_{i,i}}} - \frac{1}{C}\right)}$$
$$-\left(-\frac{1}{2}\mu^{-\frac{3}{2}}\right) \frac{\sum_{i=1}^n \left(\sqrt{\frac{A_{i,i}}{2\mu B_{i,i}}} - \frac{1}{C}\right)\sqrt{\frac{A_{i,i}}{2B_{i,i}}}}{\sum_{i=1}^n \left(\sqrt{\frac{A_{i,i}}{2\mu B_{i,i}}} - \frac{1}{C}\right)^2}.$$

For convenient representation, we define some quantities first, $\overline{x} = \sum_{i=1}^n A_{i,i}\sqrt{\frac{A_{i,i}}{2B_{i,i}}}$, $\overline{y} = \sum_{i=1}^n \frac{A_{i,i}}{2B_{i,i}}$, $\overline{z} = \frac{1}{C}\sum_{i=1}^n \sqrt{\frac{A_{i,i}}{2B_{i,i}}}$ and $\overline{u} = \frac{1}{C}\sum_{i=1}^n A_{i,i}$. By setting $\nabla_\mu g(\mu)$ to zero, and $\mu > 0$, we arrive at the following equation:

$$\frac{\overline{x}}{\frac{1}{\sqrt{\mu}}\overline{x} - \overline{u}} - \frac{\frac{1}{\sqrt{\mu}}\overline{y} - \overline{z}}{\frac{1}{\mu}\overline{y} - 2\frac{1}{\sqrt{\mu}}\overline{z} + \frac{n}{C^2}} = 0.$$

Finally, the solution of (9) is $\mu = \left(\frac{\overline{y}\overline{u} - \overline{z}\overline{x}}{\overline{z}\overline{u} - \frac{n}{C^2}\overline{x}}\right)^2$. □

We also consider to apply the similar strategy to choose the best $\mu$ for both diffusion kernel and Gaussian field kernel. However, we cannot get a closed-form solution for them.

### 4.3 Insensitivity to Model Parameters

In order to solve our proposed problem (7), we derive two update rules: one is for proposed model using (8), and another is for model selection using (10). Moreover, the following proposition shows that the final decision function of the second model is insensitive to both $\mu$ and $C$.

**Proposition 4.** *Based on Proposition 2 and 3, the proposed model in (7) is independent of parameters $\mu$ and $C$, and the decision function can be written as:*

$$f(x_u) = (\overline{K} - I)_{u,l}\overline{K}_{l,l}^{-1}\mathbf{y}_l,$$

*where $\overline{K} = U\overline{\Sigma}U^T$, $\overline{\Sigma} = diag([\overline{\lambda}_1, \ldots, \overline{\lambda}_n])$, $\overline{\lambda}_i = \left|\frac{\widehat{z}\widehat{u} - n\overline{x}}{\widehat{y}\widehat{u} - \widehat{z}\overline{x}}\right|\sqrt{\frac{A_{i,i}}{2B_{i,i}}}$, $\widehat{z} = \sum_{i=1}^n \sqrt{\frac{A_{i,i}}{2B_{i,i}}}$ and $\widehat{u} = \sum_{i=1}^n A_{i,i}$.*

*Proof.* We substitute equation (10) into equation (8), and obtain an equation without parameter $\mu$ as follows:

$$\lambda_i = \sqrt{\frac{A_{i,i}}{2\left(\frac{\overline{y}\overline{u} - \overline{z}\overline{x}}{\overline{z}\overline{u} - \frac{n}{C^2}\overline{x}}\right)^2 B_{i,i}}} - \frac{1}{C} = \left|\frac{\overline{z}\overline{u} - \frac{n}{C^2}\overline{x}}{\overline{y}\overline{u} - \overline{z}\overline{x}}\right|\sqrt{\frac{A_{i,i}}{2B_{i,i}}} - \frac{1}{C}.$$

Here, we set $\widehat{z} = \sum_{i=1}^n \sqrt{\frac{A_{i,i}}{2B_{i,i}}}$ and $\widehat{u} = \sum_{i=1}^n A_{i,i}$, then

$$\lambda_i = \frac{1}{C}\left|\frac{\widehat{z}\widehat{u} - n\overline{x}}{\widehat{y}\widehat{u} - \widehat{z}\overline{x}}\right|\sqrt{\frac{A_{i,i}}{2B_{i,i}}} - \frac{1}{C}.$$

The new kernel can be written as $K = U\Sigma U^T = \frac{1}{C}U(\overline{\Sigma} - I)U^T = \frac{1}{C}(\overline{K} - I)$, and $K + \frac{1}{C}I = \frac{1}{C}\overline{K}$ where $\overline{K} = U\overline{\Sigma}U^T \succeq 0$ since $\overline{\lambda}_i = \left|\frac{\widehat{z}\widehat{u} - n\overline{x}}{\widehat{y}\widehat{u} - \widehat{z}\overline{x}}\right|\sqrt{\frac{A_{i,i}}{2B_{i,i}}} \geq 0$. The solution for RLS with dual variables $\alpha$ can be rewritten as:

$$\alpha = (K_{l,l} + \frac{1}{C}I_l)^{-1}\mathbf{y}_l = C\overline{K}_{l,l}^{-1}\mathbf{y}_l.$$

For classification problem, we have the decision function

$$f(x_u) = K_{u,l}\alpha = (\overline{K} - I)_{u,l}\overline{K}_{l,l}^{-1}\mathbf{y}_l.$$

Table 1: Statistics of datasets: $c$ is the number of classes, $d$ is the input dimension, $n_l$ is the number of labeled examples, $n$ is the total number of examples (including labeled and unlabeled data), $k$ is the number of nearest neighbors in the graph, $p$ is the degree used in graph Laplacian matrix

| Dataset   | c  | d      | $n_l$  | n     | k   | p |
|-----------|----|--------|--------|-------|-----|---|
| G50C      | 2  | 50     | 50     | 550   | 50  | 5 |
| USPST     | 10 | 256    | 50     | 2,007 | 10  | 2 |
| PCMAC     | 2  | 7,511  | 50     | 1,946 | 50  | 5 |
| webKB(P)  | 2  | 3,000  | 12     | 1,051 | 200 | 5 |
| webKB(PL) | 2  | 4,840  | 12     | 1,051 | 200 | 5 |
| DIGIT1    | 2  | 241    | 10/100 | 1,500 | 5   | 2 |
| USPS      | 2  | 241    | 10/100 | 1,500 | 5   | 2 |
| COIL2     | 2  | 241    | 10/100 | 1,500 | 5   | 2 |
| COIL6     | 6  | 241    | 10/100 | 1,500 | 5   | 2 |
| TEXT      | 2  | 11,960 | 10/100 | 1,500 | 50  | 5 |

The decision function obtained by our proposed spectral kernel learning algorithm is not a function of parameters $C$ and $\mu$. Therefore, our proposed method does not depend on parameters $C$ and $\mu$. □

---

**Algorithm 1** SKL_KTA Algorithm

1: Build Laplacian matrix $L$
2: Do eigenvalue decomposition $[U, \Upsilon] = eig(L)$
3: Set $A = U_l^T y_l y_l^T U_l$, $B = \Upsilon$, and $B_{i,i} \leftarrow B_{i,i} + \epsilon$
4: Compute $\overline{K} = U\overline{\Sigma}U^T$ where $\overline{\lambda}_i$ is defined in Proposition 4
5: Decision function $f(x_u) = (\overline{K} - I)_{u,l}\overline{K}_{l,l}^{-1}y_l$

---

### 4.4 Algorithms

Since the graph Laplacian matrix is defined on a graph $G$, the number of eigenvalue equals to zero in $L$ is the number of connected components in the graph. There is at least one zero eigenvalue in the eigensystem of $L$ because each graph at least has one connected component as itself. So there exists $B_{i,i} = 0$, from (8), the solution is not well-defined. In order to avoid this, we add a small ridge $\epsilon$ to $V$ in our implementation. This also guarantees $\overline{K}_{l,l}$ is well-defined as well. The proposed parameter-free spectral kernel learning algorithm by learning $\mu$ using KTA, namely SKL_KTA, is shown in Algorithm 1.

---

**Algorithm 2** SKL Algorithm

1: Given $C$ and $\mu$
2: Build Laplacian matrix $L$
3: Do eigenvalue decomposition $[U, \Upsilon] = eig(L)$
4: Set $A = U_l^T y_l y_l^T U_l$, $B = \Upsilon$, and $B_{i,i} \leftarrow B_{i,i} + \epsilon$
5: Compute $K = U\Sigma^*U^T$ where $\lambda_i^*$ is defined in Proposition 2
6: Decision function $f(x_u) = K_{u,l}(K_{l,l} + \frac{1}{C}I_l)_{l,l}^{-1}y_l$

---

Table 2: Testing accuracies (in %) of different methods on all data sets. The number in parentheses shows the relative rank of the algorithm (performance-wise) on the corresponding dataset. The smaller the rank, the better the relative performance is. $n_l$ is the size of labeled data for different settings for the datasets.

| $n_l$ | Dataset | LapRLS | LapSVM | SKL_KTA | SimpleMKL | SVM |
|---|---|---|---|---|---|---|
|  | G50C | 94.64 ± 0.70 (1) | 94.26 ± 1.90 (3) | 94.60 ± 0.64 (2) | 90.40 ± 1.79 (5) | 91.08 ± 1.50 (4) |
|  | USPST | 89.44 ± 1.80 (2) | 89.20 ± 1.41 (3) | 89.67 ± 1.32 (1) | 65.86 ± 16.9 (5) | 79.33 ± 2.33 (4) |
|  | PCMAC | 90.89 ± 0.90 (1) | 89.89 ± 2.37 (3) | 89.95 ± 0.81 (2) | 86.05 ± 1.93 (4) | 78.90 ± 2.37 (5) |
|  | webKB(P) | 95.16 ± 0.55 (1) | 92.46 ± 2.73 (3) | 94.57 ± 0.60 (2) | 91.07 ± 3.51 (5) | 91.09 ± 2.67 (4) |
|  | webKB(PL) | 94.94 ± 3.84 (3) | 87.52 ± 3.13 (5) | 95.14 ± 1.32 (1) | 94.99 ± 1.09 (2) | 93.87 ± 2.11 (4) |
|  | Digit1 | 85.44 ± 9.70 (3) | 89.50 ± 6.11 (2) | 93.47 ± 5.20 (1) | 57.51 ± 10.85 (5) | 77.05 ± 5.24 (4) |
|  | USPS | 80.54 ± 1.73 (2) | 70.24 ± 14.37 (5) | 83.53 ± 2.09 (1) | 80.06 ± 0.16 (3) | 80.02 ± 0.05 (4) |
| 10 | COIL2 | 55.98 ± 8.62 (3) | 60.55 ± 8.86 (2) | 66.19 ± 7.39 (1) | 52.11 ± 3.88 (4) | 51.57 ± 3.12 (5) |
|  | COIL6 | 41.82 ± 6.24 (2) | 42.35 ± 5.50 (1) | 40.79 ± 6.79 (3) | 17.23 ± 1.29 (5) | 17.86 ± 3.02 (4) |
|  | TEXT | 61.86 ± 10.2 (1) | 60.43 ± 9.67 (2) | 58.17 ± 6.30 (3) | 55.06 ± 6.12 (4) | 53.90 ± 6.00 (5) |
|  | Digit1 | 98.14 ± 0.56 (1) | 97.37 ± 1.14 (3) | 97.80 ± 0.44 (2) | 94.79 ± 1.27 (4) | 92.23 ± 1.44 (5) |
|  | USPS | 95.10 ± 1.49 (1) | 93.38 ± 4.17 (3) | 94.36 ± 1.54 (2) | 82.76 ± 2.69 (5) | 89.86 ± 1.03 (4) |
| 100 | COIL2 | 95.75 ± 2.47 (2) | 94.88 ± 2.16 (3) | 97.39 ± 0.96 (1) | 53.24 ± 4.75 (4) | 49.68 ± 0.17 (5) |
|  | COIL6 | 87.35 ± 2.11 (1) | 86.08 ± 3.20 (3) | 86.55 ± 2.11 (2) | 21.74 ± 6.03 (4) | 16.30 ± 0.23 (5) |
|  | TEXT | 76.44 ± 1.38 (2) | 76.61 ± 1.90 (1) | 74.67 ± 1.41 (3) | 74.62 ± 2.46 (4) | 51.12 ± 4.94 (5) |
|  | Average rank | 1.73 | 2.80 | 1.80 | 4.20 | 4.47 |

We can also initialize parameters $C$ and $\mu$ in advance, and the propose spectral kernel learning algorithm without parameter learning procedure, namely SKL, is shown in Algorithm 2. SKL could employ cross-validation method to choose the parameters $C$ and $\mu$ from the input data if the model parameters are not given.

The complexities of the both proposed algorithms are the same, but SKL_KTA has no model parameter if given the graph Laplacian matrix ahead of time; while SKL still has parameters $C$ and $\mu$, which may need to be tuned by validation methods. Moreover, both algorithms can be naturally extended into multi-class problem by replacing $y_i$ with a vector $Y_i$ with only one 1 at the position indicating label $y_i$ and 0 for the rest.

## 5 Experiments

### 5.1 Datasets

Experiments were performed on various popular datasets described in Table 1. G50C is a toy dataset generated from two unit covariance normal distributions with equal probabilities. The class means are adjusted so that the Bayes error is 5%. USPST is taken from the USPS(test) dataset. The text data consists of binary classification problems: PCMAC is a two-class dataset generated from 20newsgroups collection in order to categorize documents into two topics: Macintosh and Windows systems. WebKB is a subset of web document of the computer science departments of four universities. The two categories are *course* or *non-course*. For each document, there are two representations: the textual content of web page (PAGE) and the anchor text on links on other web pages pointing to the web page(LINK). We also considered a joint(PAGE+LINK) representation by concatenating the features. DIGIT1, USPS, COIL2, COIL6 and TEXT are the benchmark datasets in (Chapelle et al., 2006). In our experiments, we use both the 10 labeled examples and the 100 labeled examples settings. For each dataset, there are twelve subsets.

### 5.2 Transductive Setting

In transductive setting, the training set comprises of $n$ examples, $n_l$ of which are labeled. The performance for each algorithm is reported by predicting the labels of the $n_u$ unlabeled examples. The experimental setup is based on (Belkin et al., 2006) and (Chapelle et al., 2006):

We compare our proposed methods with Laplacian SVM (LapSVM) and Laplacian RLS (LapRLS) (Belkin et al., 2006). For LapSVM and LapRLS, it is very time consuming to select all the parameters by cross-validation, so we first fix some parameters. The graph Laplacian $L$ is defined as $L = D - S$ where $S_{i,j} = \exp(-\frac{||x_i - x_j||^2}{2\sigma_g^2})$ if $x_i$ and $x_j$ are adjacent and zeros otherwise, and $D$ is a diagonal degree matrix given by $D_{i,i} = \sum_i S_{i,j}$. We fix parameter with the mean of distance on connected edge $\sigma_g^2 = \frac{1}{|E|} \sum_{(i,j) \in E} ||x_i - x_j||^2$ where $E$ is the set of undirected edge in the graph. The normalized graph Laplacian matrix $L = I - D^{-\frac{1}{2}} S D^{-\frac{1}{2}}$ is used in our experiments. The number of nearest neighbors $k$ and the degree of graph Laplacian matrix $p$ for all the datasets can be seen in Table 1. The setting for $k$ and $p$ can be referred to (Belkin et al., 2006) and (Chapelle et al., 2006). Gaussian kernel $k(x_i, x_j) = \exp(-\frac{||x_i - x_j||^2}{2\sigma^2})$ is used for all datasets listed in Table 1 with fixed $\sigma = \frac{1}{n} \sum_{i=1}^{n} ||x_i||$. Euclidean nearest neighbor graphs with Gaussian weights were used.

The optimal weights of the ambient and intrinsic norms, $\gamma_I$ and $\gamma_A$ were determined by varying them on the grid $\{10^{-6}, 10^{-4}, 10^{-2}, 10^{-1}, 1, 10, 10^2\}$ and chosen with respect to validation error. For each of dataset in Table 1, we constructed 10 random splits into labeled and unlabeled sets. The sizes of these sets are given in Table 1. For each split, five-fold cross-validation is used to select the best pair $\gamma_A$ and $\gamma_I$. We divide labeled data into five folds in which at least one example per class is present on each of them without additional balancing constraints. At each time, only one fold is used for labeled examples, and the rest is the validation set. All results are averaged over all splits or subsets of the available data.

Table 3: Time (in second) used in different methods on all data sets.

| $n_l$ | Dataset | Lap | LapRLS | LapSVM | SKL_KTA | SimpleMKL | SVM |
|---|---|---|---|---|---|---|---|
| | G50C | 2.5 | $235.3 \pm 106.96$ | $161.0 \pm 6.33$ | $1.8 \pm 0.50$ | $1.1 \pm 0.45$ | $0.1 \pm 0.47$ |
| | USPST | 14.1 | $15{,}296.0 \pm 203.11$ | $20{,}189.0 \pm 15.85$ | $23.3 \pm 0.83$ | $7.6 \pm 1.54$ | $1.0 \pm 0.03$ |
| | PCMAC | 52.4 | $1{,}544.4 \pm 63.59$ | $1{,}818.6 \pm 4.76$ | $18.8 \pm 0.78$ | $226.9 \pm 2.68$ | $27.4 \pm 0.47$ |
| | webKB(P) | 27.3 | $401.8 \pm 28.14$ | $449.5 \pm 7.32$ | $4.7 \pm 1.04$ | $26.7 \pm 1.29$ | $1.7 \pm 0.07$ |
| | webKB(PL) | 29.8 | $371.7 \pm 5.63$ | $450.6 \pm 6.74$ | $4.0 \pm 0.49$ | $65.6 \pm 3.96$ | $2.7 \pm 0.10$ |
| | Digit1 | 7.5 | $792.0 \pm 2.19$ | $962.5 \pm 2.71$ | $10.6 \pm 0.37$ | $0.7 \pm 0.07$ | $0.2 \pm 0.08$ |
| | USPS | 7.4 | $793.2 \pm 2.58$ | $963.1 \pm 1.96$ | $10.5 \pm 0.46$ | $1.0 \pm 0.40$ | $0.2 \pm 0.03$ |
| 10 | COIL2 | 6.8 | $795.5 \pm 3.46$ | $967.1 \pm 3.10$ | $10.1 \pm 0.23$ | $3.2 \pm 0.59$ | $0.5 \pm 0.03$ |
| | COIL6 | 7.0 | $1{,}001.7 \pm 7.71$ | $1{,}241.0 \pm 4.94$ | $10.5 \pm 1.14$ | $8.9 \pm 5.96$ | $0.2 \pm 0.01$ |
| | TEXT | 39.2 | $776.3 \pm 4.20$ | $951.2 \pm 1.70$ | $10.5 \pm 0.33$ | $444.1 \pm 47.3$ | $6.4 \pm 0.70$ |
| | Digit1 | 6.8 | $798.8 \pm 2.22$ | $964.4 \pm 2.11$ | $10.6 \pm 0.56$ | $2.4 \pm 0.32$ | $0.9 \pm 0.04$ |
| | USPS | 6.6 | $800.5 \pm 2.19$ | $966.1 \pm 2.41$ | $10.4 \pm 0.51$ | $2.5 \pm 0.40$ | $0.6 \pm 0.06$ |
| 100 | COIL2 | 7.0 | $802.6 \pm 3.65$ | $971.8 \pm 9.78$ | $10.1 \pm 0.39$ | $4.9 \pm 1.70$ | $1.3 \pm 0.04$ |
| | COIL6 | 7.1 | $4{,}781.9 \pm 21.32$ | $5{,}736.0 \pm 213.72$ | $11.0 \pm 1.22$ | $101.2 \pm 312.92$ | $1.5 \pm 0.07$ |
| | TEXT | 39.4 | $783.4 \pm 2.02$ | $951.6 \pm 1.98$ | $10.4 \pm 0.45$ | $546.1 \pm 49.88$ | $56.6 \pm 2.17$ |

### 5.3 Supervised Setting

In supervised setting, the training set only includes labeled data. It means that no unlabeled examples are used for training model. We compare our method with the state-of-the-art MKL method: SimpleMKL (Rakotomamonjy et al., 2008), and SVM using LIBSVM software (Chang and Lin, 2001), both of which only work under supervised setting.

We use the same candidate kernels as the setting in SimpleMKL (Rakotomamonjy et al., 2008) which are: (1) Gaussian kernels with 10 different bandwidths $\sigma = [0.5, 1, 2, 5, 7, 10, 12, 15, 17, 20]$, and (2) polynomial kernels of degree $1, 2, 3$.

All the candidate kernel matrices have been normalized to unit trace for SimpleMKL. For SVM, we use Gaussian kernel with default setting for $\gamma$. Both SimpleMKL and SVM have tradeoff parameter $C$. We do not perform cross-validation because the training data is too small. Therefore, we report the best average results over the 10 random splits on the grid $\{10^{-3}, 10^{-2}, 10^{-1}, 1, 10, 10^2, 10^3\}$.

### 5.4 Experimental Results

First, we compare the two proposed algorithm: SKL_KTA and SKL. We tune parameters $C$ and $\mu$ by cross-validation for SKL as described in Section 5.2. $\mu$ is on the grid $\{10^{-6}, 10^{-4}, 10^{-2}, 10^{-1}, 1, 10, 10^2\}$, and $C$ is set as described in Section 5.3. Figure 1 shows the results of the two algorithms on different datasets. For the benchmark datasets in (Chappelle, 2006), we report the results on 100 labeled data only for simplicity. We can observe that SKL_KTA is much faster than SKL while the performance does not degrade. This means that it is effective to do model selection by using KTA criterion. In the following experiments, we use SKL_KTA as our main algorithm to compare with other state-of-the-art methods.

We evaluate SKL_KTA in comparison with LapSVM, LapRLS, SimpleMKL and SVM on the datasets listed in Table 1. For multi-class problem, we use one versus rest strategy for LapSVM and LapRLS. For fair comparison, we also record all the time needed including both training with five fold cross-validation and testing on unlabeled data. Table 2 and Table 3 show the testing accuracy and time for all comparing algorithms on all datasets, respectively, and $Lap$ is the time of calculating graph Laplacian matrix. For LapRLS, LapSVM and SKL_KTA, we omit the time $Lap$ from the total time in Table 3.

As can be seen, SKL_KTA achieves competitive performance with other state-of-the-art SSL methods in term of average rank and testing accuracies. Moreover, it is better than SimpleMKL and SVM methods which only use labeled data. Table 3 indicates that the proposed method is much faster than LapSVM and LapRLS because both of them have $\lambda_I$ and $\lambda_A$ needed to be tuned by cross-validation, which is very time consuming. Even though SVM demonstrates the fastest algorithm in Table 3, SVM is implemented in C++, and ours is by MATLAB. In general, our proposed method is much cheaper than manifold regularization methods and can achieve comparable performance. We also plot the learned kernel matrix of SKL_KTA and SimpleMKL (trained from only 50 labeled patterns) on the USPST data (2,007 patterns) in Figure 2. We can observe that SKL_KTA can learn the structure of *ideal kernel* (Kwok and Tsang, 2003) for 10 classes, which explains the reason why SKL_KTA achieves the superior predictive performance over SimpleMKL.

## 6 Conclusion

In this paper, we propose two very simple semi-supervised kernel learning algorithms, namely SKL_KTA and SKL, which can be solved efficiently without using any optimiza-tion solver. Moreover, for a given graph Laplacian matrix, the model parameters of SKL_KTA can also be learned automatically by maximizing the kernel target alignment on labeled data. Experimental results on various real world datasets show that the proposed SKL algorithms can obtain comparable performance with other state-of-the-art SSL methods, and also demonstrate better predictive performance than kernel learning in supervised setting such as SimpleMKL and SVM algorithm. Another promising result is that the proposed two-stage SKL_KTA algorithm does not require to perform any model selection by cross-validation, and in turn SKL_KTA is much faster than LapRLS and LapSVM in practice.

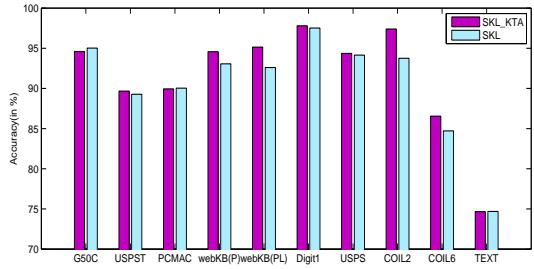 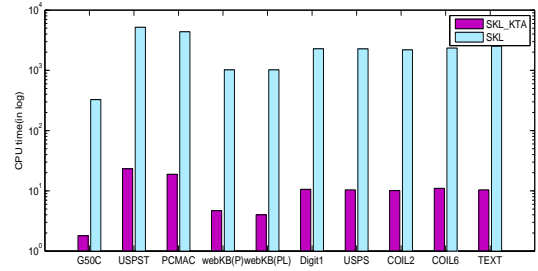

Figure 1: Testing accuracies and time of SKL and SKL_KTA on all datasets.

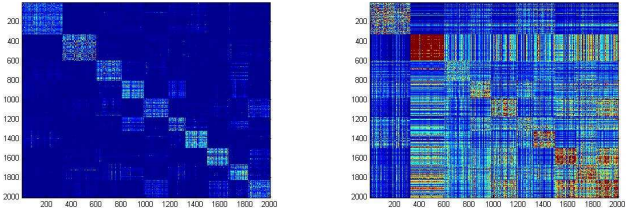

Figure 2: Learned kernel matrix of SKL_KTA (left) and SimpleMKL (right) on USPST data (10 classes).

## Acknowledgments

This research was in part supported by Singapore MOE AcRF Tier-1 Research Grant (RG15/08).

## References

F. R. Bach, G. R. G. Lanckriet, and M. I. Jordan. Multiple kernel learning, conic duality, and the SMO algorithm. In *ICML*, 2004.

M. Belkin, P. Niyogi, V. Sindhwani, and P. Bartlett. Manifold regularization: A geometric framework for learning from examples. *JMLR*, 7:2399–2434, 2006.

C.-C. Chang and C. J. Lin. *LIBSVM: a library for support vector machines*, 2001. Software available at http://www.csie.ntu.edu.tw/c̃jlin/libsvm.

O. Chapelle, J. Weston, and B. Schölkopf. Cluster kernels for semi-supervised learning. In *NIPS*, 2002.

O. Chapelle, B. Schölkopf, and A. Zien. *Semi-supervised learning*. MIT Press, 2006.

C. Cortes, M. Mohri, and A. Rostamizadeh. $l_2$ regularization for learning kernels. In *UAI*, 2009a.

C. Cortes, M. Mohri, and A. Rostamizadeh. Learning non-linear combinations of kernels. In *NIPS*, 2009b.

N. Cristianini, J. Kandola, A. Elisseeff, and J. Shawe-Taylor. On kernel target alignment. *JMLR*, 1, 2002.

L. Duan, I. W. Tsang, D. Xu, and S. J. Maybank. Domain transfer SVM for video concept detection. In *CVPR*, 2009.

S. C. H. Hoi, R. Jin, and M. R. Lyu. Learning nonparametric kernel matrices from pairwise constraints. In *ICML*, 2007.

M. Kloft, U. Brefeld, P. Sonnenburg, S. amd Laskov, K.-R. Müller, and Zien A. Efficient and accurate lp-norm multiple kernel learning. In *NIPS*, 2009.

R. I. Kondor and J. Lafferty. Diffusion kernels on graphs and other discrete structures. In *ICML*, 2002.

J. T. Kwok and I. W. Tsang. Learning with idealized kernels. In *ICML*, 2003.

G. R.G. Lanckriet, N. Cristianini, P. Bartlett, L. B. Ghaoui, and M. I. Jordan. Learning the kernel matrix with semidefinite programming. *JMLR*, 5:27–72, 2004.

W. Liu, B. Qian, J. Cui, and J. Liu. Spectral kernel learning for semi-supervised classification. In *IJCAI*, 2009.

J. S.. Nath, G. Dinesh, S. Raman, C. Bhattacharyya, A. Ben-Tal, and K. R. Tamakrishnan. On the algorithmics and applications of a mixed-norm based kernel learning formulation. In *NIPS*, 2009.

A. Rakotomamonjy, F. R. Bach, S Canu, and Y. Grandvalet. SimpleMKL. *JMLR*, 9:2491–2512, 2008.

A. J. Smola and R. Kondor. Kernels and regularization on graphs. In *COLT*, 2003.

S. Sonnenburg, G. Rätsch, C. Schäfer, and B. Schölkopf. Large scale multiple kernel learning. *JMLR*, 7:1531–1565, 2006.

J. Sun, X. Wu, S Yan, T.-S. Cheong, L.-F.and Chua, and J. Li. Hierarchical spatio-temporal context modeling for action recognition. In *CVPR*, 2009.

A. Vedaldi, V. Gulshan, M. Varma, and A. Zisserman. Multiple kernels for object detection. In *ICCV*, 2009.

D Zhou, O. Bousquet, T. N. Lal, J. Weston, and B. Schölkopf. Learning with local and global consistency. In *NIPS*, 2003.

X. Zhu, Z. Ghahramani, and J. Lafferty. Semi-supervised learning using gaussian fields and harmonic functions. In *ICML*, 2003.

X. Zhu, J. Kandola, Z. Ghahramani, and J. Lafferty. Nonparametric transforms of graph kernels for semi-supervised learning. In *NIPS*, 2004.

J. Zhuang, I. W. Tsang, and S. C. H. Hoi. SimpleNPKL: Simple non-parameteric kernel learning. In *ICML*, 2009.